\documentclass[conference]{IEEEtran}
\IEEEoverridecommandlockouts
\usepackage{cite}
\usepackage{amsmath,amssymb,amsfonts}
\usepackage{algorithmic}
\usepackage{graphicx}
\usepackage{textcomp}
\usepackage{xcolor}
\usepackage{booktabs}
\usepackage{multirow} 
\usepackage{tabularx}
\usepackage{fancybox}

\def\BibTeX{{\rm B\kern-.05em{\sc i\kern-.025em b}\kern-.08em
    T\kern-.1667em\lower.7ex\hbox{E}\kern-.125emX}}
\begin{document}

\title{Self-Evolving Human-Centered Framework for Explainable Depression Symptom Annotation}

\author{\IEEEauthorblockN{
Hoang-Loc Cao\IEEEauthorrefmark{1}\IEEEauthorrefmark{5}\IEEEauthorrefmark{3}\thanks{\IEEEauthorrefmark{1}These authors contributed equally.},
Van Pham\IEEEauthorrefmark{1}\IEEEauthorrefmark{3},
Truong Thanh Hung	Nguyen\IEEEauthorrefmark{5}\IEEEauthorrefmark{2},
Phuc Truong Loc	Nguyen\IEEEauthorrefmark{4},\\
Phuc Ho\IEEEauthorrefmark{3},
Veronica Whitford\IEEEauthorrefmark{2},
Hung Cao\IEEEauthorrefmark{2}}
\IEEEauthorblockA{
\IEEEauthorrefmark{2}University of New Brunswick, Canada\\
\IEEEauthorrefmark{3}University of Science, VNU-HCM, Vietnam, \IEEEauthorrefmark{4}Friedrich-Alexander-Universität Erlangen-Nürnberg, Germany\\
\IEEEauthorrefmark{5}Email: chloc22@clc.fitus.edu.vn, hung.ntt@unb.ca, hcao3@unb.ca}
}

\maketitle
\begin{abstract}
Annotation quality is a major bottleneck in building reliable and explainable artificial intelligence (XAI) systems for mental health research. In depression-related datasets, labels are often assigned without structured evidence, symptom-level justification, or traceable alignment to the criteria of the Diagnostic and Statistical Manual of Mental Disorders, Fifth Edition, Text Revision (DSM-5-TR), limiting both transparency and downstream model interpretability. We propose a self-evolving, expert-in-the-loop annotation framework for Major Depressive Disorder (MDD) that combines large language model (LLM)–assisted labeling with expert verification. The framework is intended to support the construction of explainable, DSM-5-TR-aligned datasets rather than to perform clinical diagnosis. It operates in three stages: candidate evidence selection from textual records, criterion-level DSM-5-TR analysis, and case-level synthesis producing label-level diagnostic and severity annotations. A dual-memory architecture, composed of Example Memory and Reflection Memory, is designed to internalize expert feedback and iteratively improve future annotations without retraining; we describe this mechanism and leave its evaluation across multiple feedback cycles to future work. In addition to final labels, the framework exports clinical evidence, reasoning traces, and edit histories, enabling comprehensive auditability. In a pilot study with expert-reviewed samples, the proposed approach improves annotation consistency and explainability while lowering manual revision effort.
\end{abstract}

\IEEEpubidadjcol
\section{Introduction}



\begin{figure*}[t]
    \centering
    \includegraphics[width=0.9\textwidth]{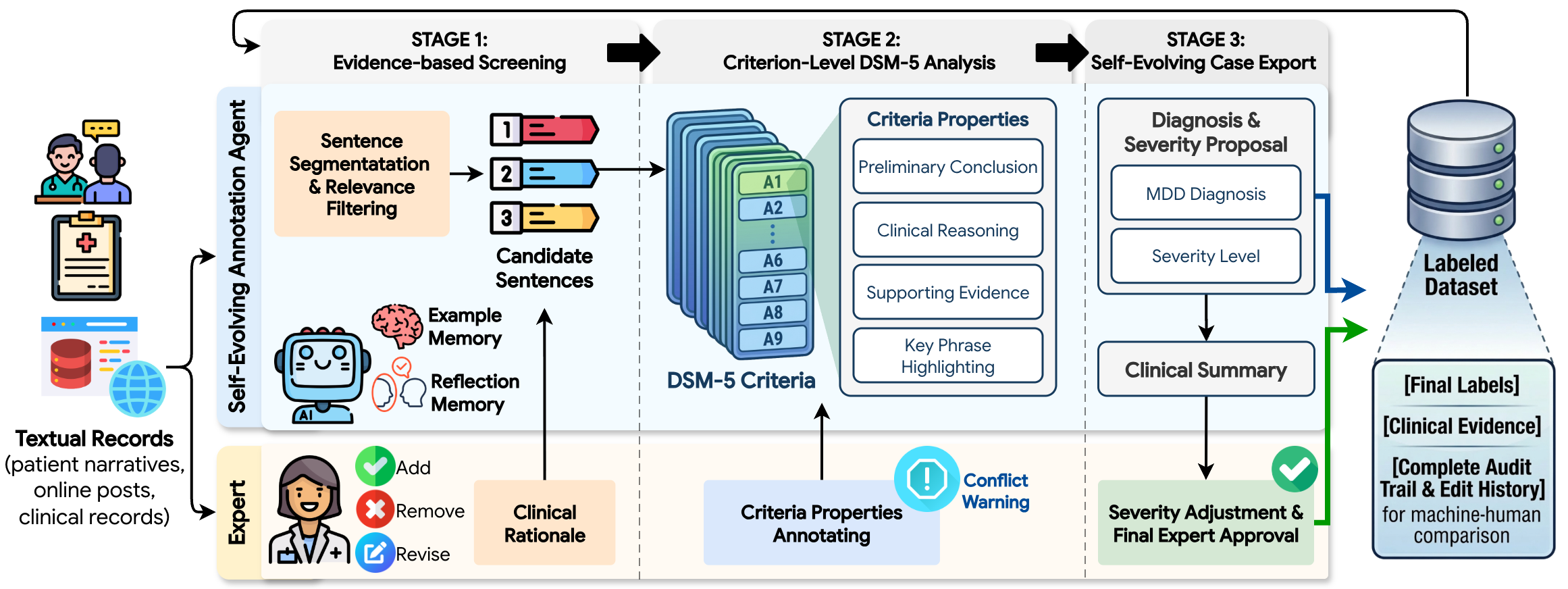}
    \caption{Overview of the proposed human-centered, self-evolving DSM-5-TR annotation framework. The pipeline consists of three stages: (1) screening and evidence selection, (2) criterion-level DSM-5-TR analysis with highlighted clinical rationale, and (3) case-level synthesis with expert validation and structured export. The framework integrates AI-assisted labeling with human verification and iterative self-evolution.}
    \label{fig:framework}
\end{figure*}

Mental health conditions contribute substantially to global health challenges, partly because their symptoms are often subjective, overlapping, context-dependent, and difficult to define consistently~\cite{WHO2025MentalDisorders}. Major depressive disorder (MDD) is especially prominent due to its prevalence, functional impact, and heterogeneous presentation~\cite{WHO2025Depression}. Rather than reflecting a single marker, MDD involves affective, cognitive, and somatic experiences, including low mood, loss of interest, irritability, hopelessness, concentration difficulties, rumination, fatigue, sleep and appetite changes, psychomotor changes, and reduced energy~\cite{Cui2024}. These experiences are documented across varied sources such as diaries, ecological momentary assessments, questionnaires, patient narratives, electronic health records, clinician notes, and informal digital communication, creating challenges for consistent interpretation even before computational modeling is introduced~\cite{Kung2021,2026-02917-001,FISHER2025104818}.
Clinical practice addresses this complexity through standardized diagnostic frameworks and validated assessment tools. The \textit{Diagnostic and Statistical Manual of Mental Disorders, Fifth Edition} (DSM-5) established a standardized diagnostic framework for classifying mental disorders using symptom-based criteria, while its \textit{Text Revision} (DSM-5-TR) updates descriptive text, terminology, coding guidance, and selected clarifications~\cite{apa2013dsm5,first2022dsm,first2023dsm}. Structured interviews such as the SCID-5 support criterion-level expert assessment~\cite{american2015structured}, and the PHQ-9 offers a brief validated measure for depression screening and severity~\cite{Kroenke2001}. These tools provide strong clinical grounding and promote consistency, but they are designed primarily for clinical assessment rather than fine-grained annotation of free text, such as evidence links, criterion-level rationales, or structured decision traces for computational reuse~\cite{Sadeghi2024}.

In parallel, artificial intelligence (AI) methods have increasingly been applied to depression-related language, particularly through natural language processing (NLP) approaches for unstructured text \cite{Fisher2026}. Their primary advantage lies in scalability: they can process large volumes of narrative data and identify patterns that may be difficult to track manually across long or heterogeneous records. However, in psychiatric applications, predictive output alone is insufficient. Clinicians, annotators, and downstream model developers must also be able to inspect the textual evidence supporting a decision, understand how that evidence relates to recognized clinical constructs, and determine whether the resulting judgment is clinically defensible. For this reason, explainable AI (XAI) is especially important in mental health settings, where transparency and interpretability are closely linked to trust, accountability, and practical usability \cite{Joyce2023,10.1145/3788686}. These considerations suggest that the most clinically useful direction is not to replace established assessment practices with opaque automation, but to integrate DSM-5/DSM-5-TR-guided clinical structure, AI-assisted scalability, and XAI-oriented transparency within a single evidence-aware annotation framework.

In this paper, we propose an Adaptive DSM-5-TR Standardized Annotation Framework for MDD prediction. We focus specifically on MDD because of its well-defined DSM-5-TR criteria, widespread clinical assessment tools, and rich expression in narrative text. The framework is designed to support multi-label annotation through DSM-5-TR-aligned label standardization, adaptive suggestion generation, and expert-in-the-loop revision. In addition to producing structured annotations, the framework enables the analysis of system behavior through expert edits and evidence-aware review. We evaluate the framework in a pilot study with expert-reviewed samples, focusing on both system label quality and the amount of manual correction required.

Our contributions are summarized as follows:
\begin{itemize}
    \item \textbf{A Human-Centered Framework for Explainable DSM-5-TR Annotation:} We propose a collaborative framework that transforms depression screening into a transparent, criterion-level process, generating structured evidence, clinical rationales, and key phrase highlighting to ensure high-fidelity explainable datasets.
    \item \textbf{A Self-Evolving Mechanism via Dual-Memory Integration:} We introduce a self-evolving architecture utilizing \textit{Example} and \textit{Reflection Memories}. This allows the agent to internalize expert feedback and ``lessons learned'' to iteratively improve diagnostic alignment without retraining the model.
    \item \textbf{A Pilot Evaluation of Annotation Quality and Expert Effort:} In a pilot study on 10 cases reviewed against gold annotations from five experts, the framework achieves high agreement with expert labels while substantially reducing manual annotation effort and time and supporting expert control through interactive conflict detection.
\end{itemize}

\section{Related Work}
\subsection{Explainability in Mental Health Prediction}

Explainability is a core requirement in mental health prediction because clinically relevant signals are often subtle, context-dependent, and difficult to interpret in isolation \cite{Joyce2023}. This is especially true for depression, where NLP models are increasingly used to infer symptoms or disorder status from language, but predictive performance alone does not show why a model reached its decision \cite{Fisher2026}. Thus, based on prior work, mental health models should do more than return accurate labels. They should support expert inspection of the evidence underlying a prediction and relate model outputs to clinically meaningful constructs \cite{Joyce2023,10.1007/978-981-96-8173-0_33}. This requirement also extends to recent large language model (LLM) approaches, which can generate explanations alongside predictions; however, these explanations still require careful evaluation of quality, reliability, and clinical usefulness before they can support expert review \cite{10.1145/3589334.3648137,10.1145/3788686,nguyen2025motion2meaning}.

Existing work addresses this problem in several ways. Post-hoc methods \cite{nguyen2021evaluation} such as SHapley Additive exPlanations (SHAP) \cite{10.5555/3295222.3295230} and Local Interpretable Model-agnostic Explanations (LIME) \cite{10.1145/2939672.2939778} are commonly used to show which features most influenced depression detection or severity prediction \cite{ahmed2025explainable}. Other studies improve clinical interpretability by predicting symptoms or questionnaire items rather than only a final label, enabling outputs to be compared more directly with established assessment practice \cite{zirikly-dredze-2022-explaining,Weber2025}. Research on clinician interaction with mental-health AI further shows that explanations are most useful when they clearly state the basis of a prediction and fit naturally into expert review \cite{kelly2025investigating}. Recent interpretable LLM studies follow the same direction by pairing predictions with generated explanations, while also showing that explanation quality must be evaluated rather than assumed \cite{10.1145/3589334.3648137}. Most existing work, however, still centers on making predictions more interpretable after the fact, leaving open how explanatory content should be organized for systematic expert review.

\subsection{Explainable Mental Health Annotation Frameworks}
Annotation frameworks for explainable mental health prediction increasingly move beyond document-level labels toward clinically structured supervision. PsySym introduced a multi-disease symptom annotation framework built from DSM-based symptom classes and clinical questionnaires, with targeted retrieval and quality-control procedures to support symptom-assisted, interpretable detection \cite{zhang-etal-2022-symptom}. DepreSym narrowed this focus to depression by constructing a sentence-level corpus aligned with 21 Beck Depression Inventory-II (BDI-II symptoms), using expert assessors and detailed relevance guidelines to link text to specific depressive markers \cite{Perez2025}. A more recent depression dataset goes one step further by annotating both depressive spans and symptom categories, enabling evaluation at the level of textual evidence and symptom alignment \cite{bolegave2025gold}. Together, these studies point to a move toward finer annotation, from symptom categories to sentence-level relevance and span-level evidence.

A parallel line of work addresses how such annotations are produced and reviewed. In clinically enriched mental-health data, LLMs have been used to assist annotation and data collection, but reliable outputs still depend on expert-defined variables and clinician oversight \cite{aich-etal-2025-using}. More general human–LLM annotation systems make the same point from a workflow perspective by combining prompt control, annotation management, automatic labeling, and selective human verification \cite{kim-etal-2024-meganno}. These systems demonstrate how labels can be generated, managed, and checked using AI, but they leave room for closer integration with clinically structured symptom data in mental health settings. The remaining gap is not whether AI can assist with annotation, but how symptom-grounded evidence annotation and collaborative AI–expert review can be unified into a single pipeline for explainable mental health prediction.

\begin{figure*}[t]
    \centering
    \shadowbox{\includegraphics[width=.8\linewidth]{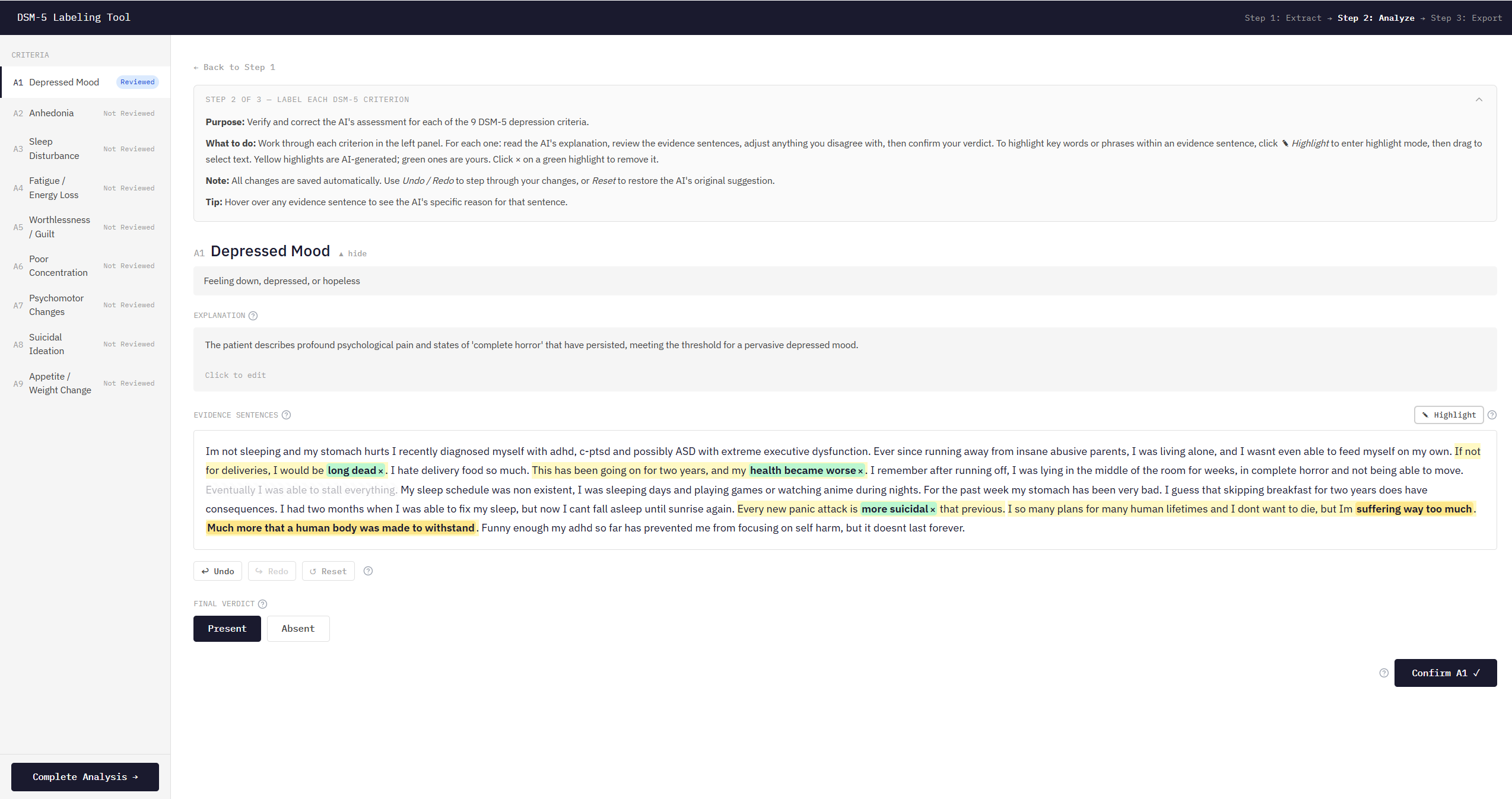}}
    \caption{Expert annotation interface for reviewing AI-suggested DSM-5-TR evidence, highlighted clinical cues, and criterion-level labels.}
    \label{fig:labeling_ui}
\end{figure*}

\section{Proposed Framework}
The proposed framework, illustrated in Fig.~\ref{fig:framework}, establishes a human-centered and self-evolving annotation ecosystem for DSM-5-TR depression labeling. Unlike traditional static pipelines that treat LLMs as independent classifiers, our approach formulates the annotation task as a collaborative, closed-loop process between a \textit{Self-Evolving Annotation Agent} and a \textit{Human Expert}.

The architecture is systematically divided into three stages: (1) Evidence-based Screening, (2) Criterion-Level DSM-5-TR Analysis, and (3) Self-Evolving Case Export. The self-evolving capability is driven by a dual-memory architecture consisting of an \textit{Example Memory} (storing gold-standard few-shot examples) and a \textit{Reflection Memory} (storing distilled clinical insights). This ensures that the agent progressively aligns with expert intuition through iterative feedback.

\subsection{Stage 1: Evidence-based Screening}
The first stage filters raw textual records (e.g., patient narratives, clinical posts) into a focused set of evidence to mitigate the expert's cognitive burden.

\subsubsection{Sentence Segmentation \& Relevance Filtering}
The agent begins by processing the input document. Utilizing contextual retrieval from the \textit{Example Memory}, the agent identifies specific segments that potentially align with depressive symptomatology. These segments are presented as a prioritized list of \textit{Candidate Sentences}.

\subsubsection{Expert Verification and Rationale}
To ensure a human-centered workflow, the \textit{Human Expert} reviews the proposed candidates. Figure~\ref{fig:labeling_ui} illustrates the expert-facing labeling interface used in this verification stage. The UI organizes the annotation process around DSM-5-TR criteria, displays AI-selected candidate evidence with highlighted clinical cues, and allows experts to accept, revise, or reject the proposed labels. This design reduces the need for exhaustive manual screening while preserving expert control over the final clinical judgment. Expert interventions are recorded as structured feedback and later used to update the framework's memory components. The expert can:

\subsection{Stage 2: Criterion-Level DSM-5-TR Analysis}
The second stage performs a granular mapping of the filtered evidence to the nine diagnostic criteria of Major Depressive Disorder (A1--A9).

\subsubsection{Criteria Properties Generation}
For each criterion, the agent generates a structured record known as \textit{criteria properties}, which includes:
\begin{itemize}
    \item \textbf{Preliminary Conclusion}: A tentative binary or categorical judgment.
    \item \textbf{Clinical Rationale:} A concise, clinically grounded explanation that links the evidence to the relevant DSM-5-TR criterion.
    \item \textbf{Supporting Evidence}: Direct quotes from the candidate sentences.
    \item \textbf{Key Phrase Highlighting}: Semantic tagging of pathognomonic terms to facilitate rapid expert verification.
\end{itemize}

\subsubsection{Criteria Properties Annotating}
The expert then performs the criteria properties annotation. This stage is augmented by a \textit{Conflict Warning} mechanism. If the agent detects contradictory clinical signals, such as symptoms suggesting comorbid anxiety or bipolar markers, it triggers a warning icon. This directs the expert's attention to ambiguous cases, ensuring that the final annotations remain clinically coherent and rigorous.

\subsection{Stage 3: Self-Evolving Case Export}
The final stage synthesizes the validated criteria into a comprehensive case profile while updating the system's internal knowledge base.

\subsubsection{Diagnosis \& Severity Proposal}
The agent aggregates the refined A1--A9 properties to generate a diagnosis and severity proposal. This module suggests an \textit{MDD Diagnosis} (applying the five-of-nine rule) and a corresponding \textit{Severity Level}. A clinical summary is then synthesized to provide a cohesive narrative of the patient’s clinical state.

\subsubsection{Final Approval and Memory Integration}
The expert performs the \textit{severity adjustment and final approval}. Once the expert signs off on the case, the framework exports a structured \textit{labeled dataset}. This dataset is uniquely comprehensive, containing:
\begin{itemize}
    \item \textbf{[Final Labels]}: High-fidelity diagnostic judgments.
    \item \textbf{[Clinical Evidence]}: Grounded text spans and highlighted phrases.
    \item \textbf{[Complete Audit Trail \& Edit History]}: A granular log of human-AI interactions for subsequent machine-human comparison.
\end{itemize}

\subsubsection{The Self-Evolution Mechanism}
The core innovation of this framework is its ability to learn from the \textit{audit trail} without retraining model parameters:
\begin{equation}
    \mathcal{K}_{t+1} = \text{Distill}(\mathcal{K}_t, \Delta_{\text{expert}}),
\end{equation}
where $\mathcal{K}$ represents the system's knowledge state and $\Delta_{\text{expert}}$ represents the delta between AI proposals and expert revisions. Expert-approved ``gold cases" populate the \textit{Example Memory}, while the distilled rationales are stored in the \textit{Reflection Memory}. This closed-loop evolution ensures that the agent’s future proposals become increasingly accurate, making the framework a truly scalable, self-improving clinical tool.

In implementation, expert revisions are converted into two memory updates. First, cases that pass final expert approval are stored as Example Memory entries containing the source text, accepted evidence spans, criterion labels, and final diagnosis. Second, recurring correction patterns are summarized into Reflection Memory entries, such as rules for distinguishing weak affective cues from DSM-5-TR A1 evidence or for rejecting unsupported evidence-to-criterion links. During subsequent annotation, the agent retrieves relevant examples and reflections using the current case representation and incorporates them into the prompting context. No model parameters are updated; adaptation occurs only through retrieval-augmented memory updates.

\begin{table*}[t]
\centering
\begin{minipage}[t]{0.57\textwidth}
\centering
\caption{Human-AI Consensus (Autonomous LLM vs Expert Gold).}
\label{tab:consensus}
\small
\setlength{\tabcolsep}{4pt}
\resizebox{\linewidth}{!}{%
\begin{tabular}{@{}lcccccccc@{}}
\toprule
\multirow{2}{*}{\textbf{Model}}
  & \multicolumn{3}{c}{\textbf{Sentence-Level}}
  & \multicolumn{3}{c}{\textbf{Criterion-Level}}
  & \textbf{Evidence}
  & \textbf{MDD Diag.} \\
\cmidrule(lr){2-4}\cmidrule(lr){5-7}
  & Prec. & Rec. & F1
  & Prec. & Rec. & F1
  & \textbf{Pair F1}
  & \textbf{Acc.} \\
\midrule
Gemini-3.5-Flash-Lite
  & \textbf{99.1} & 89.1 & \textbf{93.8}
  & 71.4 & \textbf{81.4} & 76.1
  & \textbf{67.0} & \textbf{90.0} \\[3pt]
GPT-4o-mini
  & 97.4 & 85.9 & 91.3
  & \textbf{82.9} & 79.1 & \textbf{81.0}
  & 66.7 & 80.0 \\[3pt]
GPT-5.4-mini
  & 95.9 & \textbf{91.4} & 93.6
  & 75.6 & 79.1 & 77.3
  & 57.1 & \textbf{90.0} \\
\bottomrule
\end{tabular}%
}
\end{minipage}
\hfill
\begin{minipage}[t]{0.42\textwidth}
\centering
\caption{Expert revision effort and annotation time reduction across AI-assisted labeling settings.}
\label{tab:efficiency}
\renewcommand{\arraystretch}{0.9}
\setlength{\tabcolsep}{2pt}
\small
\resizebox{\linewidth}{!}{%
\begin{tabular}{@{}lcccc@{}}
\toprule
\textbf{Modality}
  & \textbf{\shortstack{Time\\Saved (\%) $\uparrow$}}
  & \textbf{\shortstack{Total\\Edits $\downarrow$}}
  & \textbf{\shortstack{Crit.\\Flips $\downarrow$}}
  & \textbf{\shortstack{Evid.\\Edits $\downarrow$}} \\
\midrule
Expert                   &       &      &     &     \\
+ Gemini-3.5-Flash-Lite  & \textbf{75.0} & \textbf{10.2} & 2.2 & \textbf{6.5} \\
+ GPT-4o-mini            & 70.0  & 10.4 & \textbf{1.6} & 6.7 \\
+ GPT-5.4-mini           & 63.0  & 12.9 & 2.0 & 9.3 \\
\bottomrule
\end{tabular}%
}
\end{minipage}
\end{table*}

\section{Experiment and Results}

\subsection{Experimental Setup}
We conduct a pilot evaluation on 10 complex clinical cases sampled from the ReDSM5 depression-related benchmark dataset~\cite{bao2025redsm5}. Each case includes narrative evidence supporting the nine DSM-5-TR criteria for MDD. To construct expert gold labels, five experts in psychology and related fields independently annotated the cases under a shared DSM-5-TR protocol. The annotation includes clinically relevant sentence selection, criterion-level labels, evidence-to-criterion links, and final MDD diagnosis. Disagreements were cross-checked and resolved through consensus adjudication, producing gold annotations for evaluating the proposed framework.

After gold labels were established, the same cases were processed by our framework using three LLM backbones: Gemini-3.5-Flash-Lite, GPT-4o-mini, and GPT-5.4-mini. We evaluate the autonomous outputs before expert correction to assess the intrinsic quality of AI-generated annotations. The evaluation covers four levels: sentence-level evidence screening, DSM-5-TR criterion classification, evidence-pair alignment, and case-level diagnosis. In addition, we measure expert revision effort and annotation efficiency to quantify the practical utility of the human-AI workflow.

\subsection{Evaluation Metrics}
We evaluate the framework using both consensus-based and human-effort metrics. For autonomous annotation quality, we report precision, recall, and F1-score at the sentence level and DSM-5-TR criterion level, measuring how well the system identifies clinically relevant evidence and assigns symptom labels compared with expert gold annotations. We also compute evidence-pair F1, where each prediction is considered correct only if the model links the correct DSM-5-TR criterion to the correct supporting sentence. Case-level MDD diagnosis accuracy is used to measure agreement with expert final diagnoses. 

To assess human-centered utility, we report time saved, total edits, criterion flips, and evidence edits, which quantify the amount of expert correction required when using AI-generated annotations instead of manual annotation. These effort-based metrics are computed at the case level and then averaged across the 10 evaluated samples. Time saved is calculated by comparing the average manual annotation time with the average AI-assisted review time. Total edits measure the average number of expert corrections per case, while criterion flips and evidence edits separately capture changes to DSM-5-TR criterion decisions and supporting evidence links.

\subsection{Human-AI Consensus Rate}
Table~\ref{tab:consensus} shows that the proposed framework achieves strong agreement with expert gold labels across all LLM backbones. At the sentence level, all models achieve F1 scores above 91\%, indicating that the system can reliably identify clinically relevant evidence from long patient narratives. Gemini-3.5-Flash-Lite achieves the best sentence-level precision and F1 Score, with 99.1\% precision and 93.8\% F1, while GPT-5.4-mini achieves the highest recall at 91.4\%. This suggests that the framework is effective both for precise evidence filtering and broad evidence retrieval, depending on the selected backbone.

At the DSM-5-TR criterion level, GPT-4o-mini performs best, reaching 82.9\% precision and 81.0\% F1. This indicates stronger capability in mapping textual evidence to the correct symptom categories. Gemini-3.5-Flash-Lite achieves the highest criterion recall of 81.4\%, indicating greater sensitivity in detecting current symptoms. GPT-5.4-mini remains competitive, with 77.3\% criterion F1 and 90.0\% diagnosis accuracy. These results demonstrate that the system is not dependent on a single model; instead, it provides a structured DSM-5-TR reasoning layer that can operate effectively with different LLM backbones.

Evidence-pair alignment is the most challenging setting because it requires the model to identify both the correct DSM-5-TR criterion and the correct supporting sentence. Gemini-3.5-Flash-Lite achieves the highest evidence-pair F1-score at 67.0\%, closely followed by GPT-4o-mini at 66.7\%. The lower evidence-pair scores compared with sentence-level and criterion-level scores show that diagnosis prediction alone is insufficient for evaluating clinical annotation systems. A clinically useful system must also produce traceable and verifiable evidence links.

At the case level, Gemini-3.5-Flash-Lite and GPT-5.4-mini both achieve 90.0\% MDD diagnosis accuracy, while GPT-4o-mini achieves 80.0\%. This confirms that the framework can produce reliable final diagnostic decisions while preserving intermediate evidence and criterion annotations. Overall, the consensus results show that the proposed system supports multi-level clinical reasoning: it can screen evidence, classify DSM-5-TR symptoms, ground predictions in text, and produce accurate case-level diagnoses.

\subsection{Efficiency and Expert Revision Effort}
Table~\ref{tab:efficiency} evaluates how much expert effort is reduced when using the proposed framework. Compared with manual annotation, all AI-assisted settings substantially reduce annotation time. Gemini-3.5-Flash-Lite achieves the largest time saving at 75.0\%, followed by GPT-4o-mini at 70.0\% and GPT-5.4-mini at 63.0\%. This indicates that the system can shift the expert's role from fully manual annotation to efficient verification and correction of structured AI-generated drafts.

The edit-based metrics further clarify the practical annotation burden. Gemini-3.5-Flash-Lite requires the fewest total edits (10.2 per case) and the fewest evidence edits (6.5 per case). This is consistent with its highest F1-score in the strongest evidence pair in Table~\ref{tab:consensus}, suggesting that it provides the most usable evidence-grounded annotations. GPT-4o-mini requires a similar number of total edits, 10.4 per case, and achieves the fewest criterion flips, with only 1.6 flips per case. This aligns with its strongest criterion-level F1-score and indicates that GPT-4o-mini is particularly reliable for DSM-5-TR symptom classification.

GPT-5.4-mini achieves strong sentence recall and diagnosis accuracy, but it requires more expert correction, with 12.9 total edits and 9.3 evidence edits per case. This shows that high diagnosis accuracy does not necessarily imply low annotation cost. Fine-grained revision metrics are therefore essential for evaluating human-centered clinical annotation systems.

Overall, the efficiency results demonstrate that the proposed framework provides substantial practical benefit. It reduces annotation time by 63--75\% across LLM backbones while maintaining expert control through structured revision. The system does not replace clinical judgment; rather, it accelerates the annotation process by generating clinically meaningful drafts that experts can inspect, correct, and validate.

\subsection{Summary of Findings}
Our pilot results demonstrate that the proposed framework achieves strong human-AI agreement, produces explainable evidence-grounded DSM-5-TR annotations, and reduces expert annotation effort. The results also reveal complementary strengths across LLM backbones. Gemini-3.5-Flash-Lite performs best in sentence precision, evidence grounding, and efficiency; GPT-4o-mini is strongest in criterion-level classification; and GPT-5.4-mini provides high recall and diagnostic accuracy. Overall, these findings suggest that the framework is model-agnostic and support a consistent expert-in-the-loop clinical annotation workflow.
\section{Conclusion}
This paper presents a self-evolving DSM-5-TR annotation framework for explainable MDD prediction. By combining LLM-based evidence screening, criterion-level DSM-5-TR reasoning, and expert-in-the-loop verification, the framework produces structured labels, supporting evidence, and traceable revision histories. The pilot results show strong human-AI agreement across multiple annotation levels and demonstrate that AI-assisted labeling can substantially reduce expert revision time while preserving expert control. Through \textit{Example Memory} and \textit{Reflection Memory}, expert corrections are further converted into reusable feedback for improving future annotations without model retraining. Future work will extend the evaluation to larger, more diverse datasets; examine long-term improvement across feedback cycles; and adapt the framework to both other mental health conditions and finer-grained assessment tasks.

\section*{Acknowledgment}
This work was supported by NSERC Discovery Grant No RGPIN-2025-04478 and NSERC Discovery Supplement Award No DGECR-2025-00129.

\bibliographystyle{IEEEtran}
\bibliography{ref}

@misc{WHO2025MentalDisorders,
  author       = {{World Health Organization}},
  title        = {Mental disorders},
  year         = {2025},
  month        = sep,
  day          = {30},
  note         = {{WHO} Fact Sheet. Accessed: 2026-04-12}
}

@misc{WHO2025Depression,
  author       = {{World Health Organization}},
  title        = {Depressive disorder (depression)},
  year         = {2025},
  month        = aug,
  day          = {29},
  note         = {{WHO} Fact Sheet. Accessed: 2026-04-12}
}

@Article{Kung2021,
author={Kung, Benson and others},
title={Identifying subtypes of depression in clinician-annotated text: a retrospective cohort study},
journal={Scientific Reports},
year={2021},
month={Nov},
day={17},
volume={11},
number={1},
pages={22426},
issn={2045-2322}
}

@Article{2026-02917-001,
author={Collins, Amanda C. and others},
title={Semantic signals in self-reference: The detection and prediction of depressive symptoms from the daily diary entries of a sample with major depressive disorder.},
journal={Journal of Psychopathology and Clinical Science},
year={2025},
publisher={American Psychological Association},
address={US},
volume={134},
number={5},
pages={488-502},
}

@article{FISHER2025104818,
title = {Emotion rigidity in adolescents prospectively predicts future depressive symptoms assessed via self-report and clinical interview},
journal = {Behaviour Research and Therapy},
volume = {193},
pages = {104818},
year = {2025},
issn = {0005-7967},
author = {Hadar Fisher and others},
}

@Article{Cui2024,
author={Cui, Lulu and others},
title={Major depressive disorder: hypothesis, mechanism, prevention and treatment},
journal={Signal Transduction and Targeted Therapy},
year={2024},
month={Feb},
day={09},
volume={9},
number={1},
pages={30},
issn={2059-3635},
}

@book{apa2013dsm5,
  author    = {{American Psychiatric Association}},
  title     = {Diagnostic and Statistical Manual of Mental Disorders: {DSM}-5},
  edition   = {5},
  year      = {2013},
  publisher = {American Psychiatric Association},
  address   = {Arlington, VA}
}

@article{american2015structured,
  title={Structured clinical interview for DSM-5 (SCID-5)},
  author={American Psychiatric Association},
  journal={Washington, DC: American Psychiatric Association},
  year={2015}
}

@Article{Kroenke2001,
author={Kroenke, Kurt
and Spitzer, Robert L.
and Williams, Janet B. W.},
title={The PHQ-9},
journal={Journal of General Internal Medicine},
year={2001},
month={Sep},
day={01},
volume={16},
number={9},
pages={606-613},
issn={1525-1497}
}

@Article{Sadeghi2024,
author={Sadeghi, Misha
and others},
title={Harnessing multimodal approaches for depression detection using large language models and facial expressions},
journal={npj Mental Health Research},
year={2024},
month={Dec},
day={23},
volume={3},
number={1},
pages={66},
issn={2731-4251}
}

@Article{Fisher2026,
author={Fisher, Hadar
and others},
title={Language-based detection of depression with machine learning: systematic review and meta-analysis},
journal={npj Digital Medicine},
year={2026},
month={Feb},
day={24},
volume={9},
number={1},
pages={273},
issn={2398-6352},
}

@Article{Joyce2023,
author={Joyce, Dan W.
and others},
title={Explainable artificial intelligence for mental health through transparency and interpretability for understandability},
journal={npj Digital Medicine},
year={2023},
day={18},
volume={6},
number={1},
pages={6},
issn={2398-6352},
}

@article{10.1145/3788686,
author = {Nguyen, Hung and others},
title = {Heart2Mind: Human-Centered Contestable Psychiatric Disorder Prediction System Using Wearable ECG Monitors},
year = {2026},
publisher = {Association for Computing Machinery},
address = {New York, NY, USA},
journal = {ACM Trans. Comput. Healthcare},
month = jan,
}

@inproceedings{nguyen2021evaluation,
  title={Evaluation of explainable artificial intelligence: Shap, lime, and cam},
  author={Nguyen, Hung Truong Thanh and others},
  booktitle={Proceedings of the FPT AI Conference},
  pages={1--6},
  year={2021}
}

@article{ahmed2025explainable,
  title={Explainable ai for depression detection and severity classification from activity data: Development and evaluation study of an interpretable framework},
  author={Ahmed, Iftikhar and others},
  journal={JMIR Mental Health},
  volume={12},
  number={1},
  pages={e72038},
  year={2025},
  publisher={JMIR Publications Inc., Toronto, Canada}
}

@inproceedings{10.1007/978-981-96-8173-0_33,
author={Nguyen, Truong Thanh Hung and others},
title={Human-Centered Explainable Psychiatric Disorder Diagnosis System Using Wearable ECG Monitors},
booktitle={Advances in Knowledge Discovery and Data Mining},
year="2025",
publisher="Springer Nature Singapore",
pages="418--429",
isbn="978-981-96-8173-0"
}

@inproceedings{10.1145/3589334.3648137,
author = {Yang, Kailai and others},
title = {MentaLLaMA: Interpretable Mental Health Analysis on Social Media with Large Language Models},
year = {2024},
isbn = {9798400701719},
publisher = {Association for Computing Machinery},
address = {New York, NY, USA},
booktitle = {Proceedings of the ACM Web Conference 2024},
pages = {4489–4500},
numpages = {12},
location = {Singapore, Singapore},
series = {WWW '24}
}

@inproceedings{nguyen2025motion2meaning,
  title={Motion2Meaning: A Clinician-Centered Framework for Contestable LLM in Parkinson's Disease Gait Interpretation},
  author={Nguyen, Loc Phuc Truong and others},
    booktitle = {9th International Symposium on Chatbots and Human-centred AI (CONVERSATIONS) 2025},
    year = {2025}
}

@Article{Weber2025,
author={Weber, Samantha and others},
title={Using a fine-tuned large language model for symptom-based depression evaluation},
journal={npj Digital Medicine},
year={2025},
month={Oct},
day={07},
volume={8},
number={1},
pages={598},
issn={2398-6352}
}

@article{kelly2025investigating,
  title={Investigating How Clinicians Form Trust in an AI-Based Mental Health Model: Qualitative Case Study},
  author={Kelly, Anthony and others},
  journal={JMIR Human Factors},
  volume={12},
  number={1},
  pages={e79658},
  year={2025},
  publisher={JMIR Publications Inc., Toronto, Canada}
}

@inproceedings{zirikly-dredze-2022-explaining,
    title = "Explaining Models of Mental Health via Clinically Grounded Auxiliary Tasks",
    author = "Zirikly, Ayah  and
      Dredze, Mark",
    booktitle = "Proceedings of the Eighth Workshop on Computational Linguistics and Clinical Psychology",
    month = jul,
    year = "2022",
    address = "Seattle, USA",
    publisher = "Association for Computational Linguistics",
    pages = "30--39",
}

@inproceedings{10.5555/3295222.3295230,
author = {Lundberg, Scott M. and Lee, Su-In},
title = {A unified approach to interpreting model predictions},
year = {2017},
isbn = {9781510860964},
publisher = {Curran Associates Inc.},
address = {Red Hook, NY, USA},
booktitle = {Proceedings of the 31st International Conference on Neural Information Processing Systems},
pages = {4768–4777},
numpages = {10},
location = {Long Beach, California, USA},
series = {NIPS'17}
}

@inproceedings{10.1145/2939672.2939778,
author = {Ribeiro, Marco Tulio and Singh, Sameer and Guestrin, Carlos},
title = {"Why Should I Trust You?": Explaining the Predictions of Any Classifier},
year = {2016},
isbn = {9781450342322},
booktitle = {Proceedings of the 22nd ACM SIGKDD International Conference on Knowledge Discovery and Data Mining},
pages = {1135–1144},
series = {KDD '16}
}

@article{bolegave2025gold,
  title={A gold standard dataset and evaluation framework for depression detection and explanation in social media using llms},
  author={Bolegave, Prajval and Bhattacharya, Pushpak},
  journal={arXiv preprint arXiv:2507.19899},
  year={2025}
}

@Article{Perez2025,
author={P{\'e}rez, Anxo and others},
title={DepreSym: A Depression Symptom Annotated Corpus and the Role of Large Language Models as Assessors of Psychological Markers},
journal={Language Resources and Evaluation},
year={2025},
month={Sep},
day={01},
volume={59},
number={3},
pages={2737-2762},
issn={1574-0218},
}

@inproceedings{zhang-etal-2022-symptom,
    title = "Symptom Identification for Interpretable Detection of Multiple Mental Disorders on Social Media",
    author = "Zhang, Zhiling and others",
    booktitle = "Proceedings of the 2022 Conference on Empirical Methods in Natural Language Processing",
    year = "2022",
    address = "Abu Dhabi, United Arab Emirates",
    publisher = "Association for Computational Linguistics",
    pages = "9970--9985",
}

@inproceedings{aich-etal-2025-using,
    title = "Using {LLM}s to Aid Annotation and Collection of Clinically-Enriched Data in Bipolar Disorder and Schizophrenia",
    author = "Aich, Ankit and others",
    booktitle = "Proceedings of the 10th Workshop on Computational Linguistics and Clinical Psychology (CLPsych 2025)",
    year = "2025",
    pages = "181--192",
    ISBN = "979-8-89176-226-8",
}

@inproceedings{kim-etal-2024-meganno,
    title = "{MEGA}nno+: A Human-{LLM} Collaborative Annotation System",
    author = "Kim, Hannah and others",
    booktitle = "Proceedings of the 18th Conference of the European Chapter of the Association for Computational Linguistics: System Demonstrations",
    year = "2024",
    pages = "168--176",
}

@inproceedings{bao2025redsm5,
  title={ReDSM5: A Reddit Dataset for DSM-5 Depression Detection},
  author={Bao, Eliseo and P{\'e}rez, Anxo and Parapar, Javier},
  booktitle={Proceedings of the 34th ACM International Conference on Information and Knowledge Management},
  pages={6323--6327},
  year={2025}
}

@article{first2022dsm,
  title={DSM-5-TR: Overview of what’s new and what’s changed},
  author={First, Michael B and others},
  journal={World Psychiatry},
  volume={21},
  number={2},
  pages={218},
  year={2022}
}

@article{first2023dsm,
  title={DSM-5-TR: rationale, process, and overview of changes},
  author={First, Michael B and others},
  journal={Psychiatric Services},
  volume={74},
  number={8},
  pages={869--875},
  year={2023},
  publisher={American Psychiatric Association Washington, DC}
}

\end{document}